\documentclass[aps,twocolumn,twoside,floatfix,nofootinbib,a4paper]{revtex4}
\usepackage{graphicx}
\usepackage{marginnote}

\usepackage{hyperref}
\usepackage{enumitem,kantlipsum}
\usepackage[usenames,dvipsnames]{color}
\usepackage{amsmath}
\usepackage{amssymb}
\usepackage{amsfonts}
\usepackage{mathrsfs}
\usepackage{bm}
\usepackage[all]{xy}
\usepackage{hyperref}
\usepackage{tikz-cd}

\newcommand{\beq}{\begin{equation}}
\newcommand{\eeq}{\end{equation}}
\newcommand{\bea}{\begin{eqnarray}}
\newcommand{\eea}{\end{eqnarray}}
	
\newcommand{\ea}{\end{align*}}

\newcommand{\bma}{\begin{pmatrix}}
\newcommand{\ema}{\end{pmatrix}}

\usepackage[percent]{overpic}
\usepackage{overpic}

\begin{document}
\title{Deep neural networks from the perspective of ergodic theory}
\author{Fan Zhang} 
\affiliation{Gravitational Wave and Cosmology Laboratory, Department of Astronomy, Beijing Normal University, Beijing 100875, China}
\affiliation{Advanced Institute of Natural Sciences, Beijing Normal University at Zhuhai 519087, China}

\date{\today}

\begin{abstract}
\begin{center}
\begin{minipage}[c]{0.9\textwidth}
The design of deep neural networks remains somewhat of an art rather than precise science. By tentatively adopting ergodic theory considerations on top of viewing the network as the time evolution of a dynamical system, with each layer corresponding to a temporal instance, we show that some rules of thumb, which might otherwise appear mysterious, can be attributed heuristics. 
 \end{minipage} 
 \end{center}
  \end{abstract}
\maketitle

\raggedbottom
\section{Introduction and motivation \label{sec:Heuristics}}
Artificial neural networks have demonstrated great potential in their ability to learn existing knowledge, and interpolate or even slightly extrapolate to new situations. They however lack the ability to understand causations and other logical relations that indicate general intelligence. No matter, much of human activities are experience-based, and so the present incarnation of artificial intelligence algorithms are sufficient to revolutionize society. In particular, one can highlight medicine as one area where vast amounts experiential enigmas persist, and a doctor's effectiveness is largely restricted by their capacity to learn rather than the ability to understand. 

Furthermore, human bodies are complex systems whereby heterogeneity creates self-organization, which an artificial neural network, as also a complex dynamical system, may be able to (intentionally or unintentionally) emulate. Thereby, a network's learning power may well be tunable to arise out of emergent phenomena that share traits with the processes underlying human body functions. In this way, a network could serve as a crude simulation, and subsequently stands a better chance at accurately and efficiently learning and storing medical knowledge, providing better interpolations and extrapolations than the usually naive and linear manner with which humans try to carry out such tasks.  

Despite such great potential and practical successes, a theory of deep learning is still lacking however, so we are in want of general guidelines as to what architecture might perform well. In such a pursuit, it is beneficial to be able to examine deep neural networks from as many perspectives as possible, thereby acquiring a more complete picture. In this brief note, we advocate also adopting the ergodic theory approach by showing how it could offer simple intuitive (although arguably hand-wavy at its present state of deployment) explanations to some properties that we have observed in the behavior of the deep neural networks. We begin by relating some understood aspects of the networks to ergodicity concepts, which could then serve as the conduits connecting the two disciplines. 

\subsection{As fitting functions to training data\label{sec:fitting}}
We are just now beginning to understand some aspects of the behavior of deep neural networks. For example, one's experiences with lower dimensional non-convex optimization problems might suggest that there would be an enormous number of local minima, on the cost function surface in parameter space, that trap our optimization procedure and keep it away from the optimal solution. This is in fact not a problem \cite{2014arXiv1405.4604P}, because the local minima are replaced with saddle points in very high dimensions. The Hessian at critical points have many eigenvalues in high parameter dimensions, and it is more likely that they take on both positive and negative values giving us saddle points, so we are less likely to be trapped, and could instead slip out along the downwardly curving directions. But more fundamental than not being trapped, is that a good-enough solution is at all on that surface to begin with. In other words, \newline $\mathcal{C}_1:$ \emph{the network needs to be sufficiently flexible so the surface extends a large range of cost values.} 

Another aspect of neural networks that people had expected to be a problem is statistical sample complexity, i.e., having too many parameters, even more than the training data sets, should lead to overfitting, so while fitting to the (possibly noisy) training data may be perfect, interpolation, let alone extrapolation, would not work as the overfitted function may oscillate wildly in-between the training data points on which they are pinned. This behavior is not observed in reality.  One explanation is proposed by \cite{2020PNAS..11730063B}, which rigorously proved that having many parameters being unimportant for the fitting quality is key to the effectiveness of overfitted linear regression. An intuitive understanding being offered in literature (see e.g., \cite{2020PNAS..11730033S}) is that there are then many different solutions that work rather well, differing only in the unimportant parameters, thus it becomes much easier to find them. However, we note that for any fitting problem, one could always introduce completely spurious parameters that don't appear in the actual fitting procedure (we just vacuously append them to the formal parameter set to be fitted), thus are unimportant to the extreme, but this should not change at all the quality of output of that fitting procedure. So this explanation would make sense only if the original procedure is not overfitting to begin with, or in other words, so many of the fitting parameters in the linear regression are unimportant or spurious that the surviving important ones are so few in numbers that there is no overfitting in reality.
This is thus a rather trivial explanation -- the overfitting problem is not problematic if for the problem at hand the fitting architecture is nothing but superficially over-parameterized. This does not appear to be the case for deep neural networks though, for changing the network structure, especially the number of neurons, appears to alter the fitting quality. Therefore, an alternative explanation for why overfitting does not present a catastrophic snag may be needed (see end of Sec.~\ref{sec:conclusion} for further discussions). 

To this end, we first note that condition $\mathcal{C}^1$ already helps, as overfitting often occurs in regression problems when the fitting curve is just sufficiently flexible to hit all the training data points if we strong-arm it, but not enough to do so smoothly if noises are present or the choice of basis functions are not appropriate. This rigidity means the fitting curves have to take a large detour in order to make the correct directional changes in order to hit the next training point needing to be fitted, much like how a fast aircraft needs to make a very large divagation in order to make a turn. When we do interpolation or extrapolation, we have to sample from within those large detours and thus yield terrible results. When the parameter number goes much larger than training point population however, the curve becomes extremely flexible, so it could possibly effortlessly (smoothly) transit between training data points, without having to take up some very contorted shape that shoots off to large extremes in the intervening values. This is the case with deep neural networks that are complex and flexible (see e.g.,  \cite{HORNIK1991251,2019arXiv190102220E} for neutral networks as universal approximators). This is not enough though. If the fitting quality to the training data is the only criteria in the cost function, then there is in principle no guarantee that the flexible fitting curve won't still whip around like crazy. That is, the smooth curves exist, but are not necessarily the ones we get if we do not deliberately look for them. To cure this, one typically introduces these so-called regularizers into the cost functions to penalize undesired parameters that could cause instabilities of the output, as one rather contrived approach to achieve  \newline $\mathcal{C}_2:$ \emph{the network needs to be sufficiently wellposed so the output doesn't depend so extremely sensitively on initial data that infinitesimally separated initial data points leads to wildly diverging outcomes.}

\subsection{As dynamical systems}
The conditions $\mathcal{C}_1$ and $\mathcal{C}_2$ are qualitative and difficult to translate into some enforceable quantitative criteria on network design. To make further progress, we enlist the dynamical systems view of neural networks  (see e.g., \cite{2021PhRvP..15c4092F}), whereby each layer of neurons corresponds to a time instance of a dynamical system, so updates propagating through the network according to some given initial data behaves like a time evolution of the system within a state space having the same dimension as the number of neurons in each layer, where the number of layers becomes the number of discrete\footnote{Previous literature would typically go to the continuous limit and adopt differential equation results in order to glean some insight into the behavior and properties of the neural networks (see in particular \cite{2022IJBC...3250218B}, where some criteria for neural ODEs to develop chaotic behavior is worked out). This strategy is not suitable for us however, since the chaotic nature or lack thereof, differs quite drastically for discrete and continuous systems, not least because the requirement for no-self-intersection or uniqueness of trajectories is so very much relaxed for discrete evolutions. The (spatial) differentiability also effectively restricts $K_{\alpha}^{[j] \beta}$ of Eq.~\eqref{eq:LayerDetails} below to tridiagonal or other nearly diagonal (depending on the order of derivatives and the finite difference scheme adopted) forms that limit the dependence of a neuron's time evolution to only its nearest neighbours.} time steps. This dynamical systems perspective had already yielded important insight, e.g, the Lemma 1 of \cite{2018InvPr..34a4004H},  which is a well-established result applicable to the numerical methods for solving differential equation, essentially imposes a necessary condition for $\mathcal{C}_2$. On the other hand, the same paper also recognizes the importance for the forward propagation to not be too lossy (i.e., the Lyapunov exponents cannot all be too negative, or the state space volume shrinks very quickly, losing the ability to distinguish information in the initial data), essentially offering a caution on what might definitely violate $\mathcal{C}_1$. 

In this note, we further enlist the powerful mathematical tool of ergodic theory\footnote{The usual ergodic theory studies measure-invariant or state space volume preserving evolutions, or in other words, there is no dissipation in the system, which may not be the case for deep neural networks. But dissipation may not matter directly, because the sum of \emph{all} Lyapunov exponents determines whether we have dissipation, but chaos ($\exists$ at least one positive exponent), ergodicity, weakly-mixing (only constant functions are the eigenfunctions corresponding to the exponents as eigenvalues) and entropy (sum of all positive exponents) are really more about the positive ones. Also, as discussed in the main text in the last paragraph, a well-designed neural network should not be too dissipative.} to help tighten the discussion, inching a little more towards necessary and sufficient conditions. In this language, $\mathcal{C}_1$ translates to a desirability to have ergodicity\footnote{No invariant subsets that trap orbits thus prevent effective migration in e.g., the classification problem. Beginning from any set of initial data, the bundle of trajectories will eventually cover almost all of the allowed state space (besides perhaps sets of zero measure; these don't matter for statistical considerations, but if we are aiming to train the neural network to search for very special properties that nearly never occurs in a big data set, these will be relevant and so neural networks cannot produce master pieces of exceptional qualities, they are however efficient at churning out mediocrity). We can see this from the celebrated Birkhoff's ergodicity theorem that underlies statistical mechanics, which states that spatial average equals temporal average along the dynamical evolution, so any macrostate with non-vanishing measure must eventually be reached.}, while $\mathcal{C}_2$ means we should avoid mixing\footnote{\label{fn:mixing}Stronger than ergodicity, requiring that memory of the initial data be completely lost as we take many steps, in the sense that the trajectories emerging from any initial data set will have to spread out over the entire state space completely randomly according to the probability distribution of the overall state space, with conditional probability conditioning on the initial data adding no further information. For mixing systems, trajectories from any two initially disjoint initial data sets get meshed together thoroughly and permanently after sufficiently long evolution. Note mere ergodicity could also mesh two bundles of trajectories out of two disjoint sets of initial data, but the meshing can be transitory, meaning the two bundles of trajectories can separate again, and then repeat in the meshing and separating cycle (cf., the claim that merely ergodic systems are not even apparently random \cite{sep-ergodic-hierarchy}). With mixing though, the meshing is permanent with no subsequent re-separation. In a way, one can visualize the two bundles as meeting (if they meet at all) transversely over and over again with an ergodic-but-not-mixing system, but they intersect and merge tangentially with a mixing system.}. This preference to wedge in-between ergodicity and mixing is easiest to understand in regard to classification tasks, where we need ergodicity to be able to move any initial point almost anywhere else in state space to achieve segregation (see e.g., Fig.~3 in \cite{2021PhRvP..15c4092F}), but not mixing that causes neighbourhoods to all get mangled together that there is no segregation possible, and classification ends up having to be done point-wise while interpolation becomes impossible.

We will hereafter refer to this dedicate balanced wildness of the dynamical system's orbits as being on the edge of chaos, taking the view that strong mixing is a hallmark for chaotic systems\footnote{Even though it does not necessarily imply exponential divergence between trajectories, so mixing systems are sometimes referred to as being weakly chaotic.} \cite{doi:10.1093/bjps/axn053,sep-ergodic-hierarchy}, while merely ergodic systems are usually not chaotic. We caution that this relationship between chaos and ergodicity concepts is not rigorous, and is but a pragmatism that is useful for practical applications. We will adopt it in this note understanding this lack of rigor. We will similarly be loose with terminologies in the interest of brevity, especially when migrating concepts defined for infinite time system to finite ones, and those defined for measure-preserving systems to more general cases.

\section{Network spectroscopy}
As always with dynamical systems, turning to spectral considerations tends to simplify computations. The relevant quantities for our considerations, in the case of deep neural networks of finite depth, are the finite time Lyapunov exponents, which measure the tendency of the dynamics to locally drive nearby trajectories apart (note the qualifying ``finite time'' does not mean they are gauges of the cumulative divergence across the entire network depth, they are still ``per layer'' quantities as the division by $[j]$ in Eq.~\ref{eq:FTLE} below shows), and can be seen as the average local Lyapunov exponents across the available depth of the network. Once we have these numbers, the various dynamical system qualities can be assessed, e.g., whether the system is dissipative (volume preserving in the Liouville sense) depends on whether the sum of the exponents is negative, and particularly relevant for us, the system is chaotic and likely practically mixing if there exists at least one positive exponent (becomes hyperchaotic if there are more than one). 

To this end, we first recast the neural network into the dynamical systems language (see e.g., \cite{2019arXiv190405657B}). For definitiveness, we assume a basic multi-layer perceptron style network. If we see any particular configuration of a layer (labelled by $[j]\in 0, \cdots N-1$) in the deep neural network as being a point in a state variable space, with each neuron within occupying a dimension (labelled by $\alpha$), then the value being carried by that neuron $y^{[j] \alpha}_i$, as corresponding to some input state $y^{[0] \beta}_i$ ($i$ indexes the training set), gives the $\alpha$th coordinate of that point in the state variable space. We can subsequently regard the layers as time steps in an evolution in this state space, where the discrete evolution is written as 
\begin{align} 
y^{[j+1] \alpha}_i 
= & \tilde{f}^{[j]\alpha}\left(y^{[j]\beta}_i\right) \label{eq:dyno}\\
=& y^{[j] \alpha}_i + f^{\alpha}\left(y^{[j]\beta}_i,{\bf u}^{[j]}\right) \Delta t\,, 
\end{align}
where usually 
\bea 
f^{\alpha}\left(y^{[j]\beta}_i,{\bf u}^{[j]}\right) = \sigma_{\beta}^{\alpha}\left(K_{\delta}^{[j] \beta} y_i^{[j]\delta}+\xi^{[j]\beta}\right)\,, \label{eq:LayerDetails}
\eea
with ${\bf u}^{[j]}$ representing the weights $K_{\alpha}^{[j] \beta}$ and bias $\xi^{[j]\alpha}$ connecting the $[j]$th layer to the $[j+1]$th (note these parameters become independent of the input once training is complete). The $\sigma_{\beta}^{\alpha}$ in Eq.~\eqref{eq:LayerDetails} is an activation function for the neurons that's usually of a diagonal form. The expression \eqref{eq:dyno} is a more abstract representation of the dynamical process, which absorbs the layer dependent variations in the parameters ${\bf u}^{[j]}$ into the $[j]$ label of $\tilde{f}^{[j]\alpha}$. 

Assuming for simplicity that the width of the layers do not change, then one obtains, for each reference trajectory (e.g., starting from the input of a training set labelled by $i$), and an end time $j$ (usually $N-1$ but we keep the definition general here; we always fix the initial time at $j=0$ however), a square matrix 
\bea
\mathbb{M}^{[j]\alpha}_{i\, \beta} \equiv \frac{\partial y_i^{[j]\alpha}}{\partial y_i^{[0]\beta}}\,.
\eea
This matrix, when acting on a perturbation vector of the initial data $\delta y^{[0]\beta}_i$, yields the leading order changes in the state $y^{[j]\alpha}_i$ at the $[j]$th layer. Therefore the exponential rate of growth of the perturbation would be related to the logarithm of its eigenvalues, rescaled by the number of time steps $[j]$. However, there is no guarantee that $\mathbb{M}^{[j]\alpha}_{i\, \beta}$ is diagonalizable or even square in the more general cases, so one instead goes to the singular values $\mu^{[j]\alpha}_{i}$ of $\mathbb{M}^{[j]\alpha}_{i\, \beta}$ in its singular value decomposition. The finite time Lyapunov exponents are then defined as\footnote{The singular values $\mu^{[j]\alpha}_{i}$ can always be chosen to be positive, since multiplying a negative singular value and the corresponding left singular vector both by minus one will preserve the validity of the decomposition.}
\bea 
\lambda^{[j]\alpha}_{i} \equiv \frac{\ln \mu^{[j]\alpha}_{i}}{[j]}\,.  \label{eq:FTLE}
\eea 

To compute such exponents, we begin with the most general expression \eqref{eq:dyno}, which after iterations yield
\begin{align} \label{eq:dyno2}
y^{[j] \alpha}_i = \circ^{j-1}_{q=0} \tilde{f}^{[q]\alpha}\left(y^{[0]\beta}_i\right)\,.
\end{align}
So when the $\tilde{f}^{[q]\alpha}$ are all differentiable (because of the need to back-propagate errors in neural networks, its derivatives usually exist, maybe aside from at isolated points like with the ReLU activation function), we can write 
\bea
\frac{\partial y_i^{[j]\alpha}}{\partial y_i^{[0]\beta}} = \delta^{\alpha}{}_{\gamma_{j}} \delta_{\beta}{}^{\gamma_{0}}\prod^{j-1}_{q=0} \tilde{J}^{[q]\gamma_{q+1}}_{i \gamma_{q}}\,, \label{eq:ProductRule}
\eea
where the local Jacobians are
\bea
\tilde{J}^{[q]\gamma_{q+1}}_{i \gamma_{q}} \equiv \frac{\partial \tilde{f}^{[q]\gamma_{q+1}}}{\partial y^{[q]\gamma_q}_i}\,. 
\eea 

If each layer is identical so all the Jacobians are the same, then the method of powers for singular value decomposition suggests that $\mathbb{M}^{[j]\alpha}_{i\, \beta}$ is essentially rank one, dominated by the largest singular value when the neural network is deep. 
This low rank scenario has minimal expressivity, so it is important that the Jacobians are varied. In this case, there is no simple relationship between the singular values of the individual Jacobians and the final $\mathbb{M}^{[j]\alpha}_{i\, \beta}$. At best, it is possible to approximately almost-diagonalize $\mathbb{M}^{[j]\alpha}_{i\, \beta}$  (taking it to a bidiagonal form) with a Householder procedure that turns all the Jacobians upper triangular \cite{doi:10.1137/S0895479897325578}. Then, the bidiagonal form is further fully diagonalized according to the ``chasing the bulge'' sweep of e.g. \cite{doi:10.1137/0911052}, using 2-D rotations that suppress the superdiagonal entries. This entire procedure is rather opaque and blends the entries of the Jacobians in nontrivial manners, therefore,  
without specializing to specific neural networks, we cannot make definitive statements regarding the finite time Lyapunov exponents. Nevertheless, we could perhaps make some vaguely probabilistic statements on how various aspects of the network architecture (e.g., the number of layers, the width of each layer, the rank of each Jacobian etc) would \emph{likely} affect the singular values of $\mathbb{M}^{[j]\alpha}_{i\, \beta}$. 

\subsection{Effect of network architectural traits}

\subsubsection{Depth of network vs.~activation function} 
A major hurdle in applying ergodic theory to deep neural networks is that the concepts of the former are defined in the infinite evolution time limit, or in other words asymptotically, while the actual networks tend to be of finite depth. For the same local spectral characteristics, the ergodicity (or transitivity in the language of topological dynamics) and mixing properties therefore manifest more fully in deeper networks. 
As a result, greater depth is desirable for networks whose individual layers don't tend to push nearby trajectories apart, i.e., for dynamics that are on the regular side, meaning barely ergodic and far from mixing. On the other hand, for networks whose individual-layer-driven local dynamics have a strong tendency to cause neighboring trajectories to diverge, or in other words, a system that is highly chaotic and deeply in the mixing regime asymptotically, shallower networks are beneficial, as transitivity may have had time to transpire while mixing hasn't (viewing these concepts through the lens of losing dependence on initial data, then mixing is a more complete amnesia that tends to happen further into the evolution than ergodicity, which is only partial loss of memory). 

With the prevailing network designs, because the activation functions (some of which are really just distributions) are more or less binary (switching between active or dormant states) by definition, the outputs of each neuron for nearby trajectories that happen to lay on either side of the threshold tend to be very different\footnote{This variability against initial data is a signature of the nonlinearity introduced by the activation functions. A truly linear activation function will lead to a constant $\mathbb{M}$ matrix, and the entire network becomes a linear regression, which may not be able to fit the training data (e.g., when two inputs in the training set are related by a simple $1/2$ rescaling, but their corresponding outputs differ not by the same rescaling). On an intuitive level, such variability across finite and not infinitesimal shifts in initial data may also be conducive to generate the ergodic-but-not-mixing behavior discussed in footnote \ref{fn:mixing}, because different bundles of trajectories experiencing different local Jacobians tends to drive the bundles, but not necessarily the individual trajectories within each bundle, to move apart.}. So we usually have the latter case, thus shallower networks may often be desirable (see e.g., \cite{10.1007/978-3-319-50127-7_46}). However, the extent to which this is true is dependent on the activation functions, which is supposed to bring in nonlinearity, thus should naturally relate to how chaotic the network evolution is. More specifically:
\begin{itemize}
\item
One category consists of the binary step, sigmoid, or tanh activation functions, which all have small derivatives far from the transition region, but a large derivative within. As a result, their Jacobians contain a delta-function style spike that, at occasions, contributes to a very large Frobenius norm to $\tilde{J}$ (we henceforth suppress unimportant indices for brevity) and subsequently $\mathbb{M}$ (sans chance cancellations). Because the Frobenius norm of $\mathbb{M}$ is the L2 norm of its singular values, this then implies large singular values, and thus large (more positive) finite time Lyapunov exponents, and consequently deeper protrusion into the mixing regime. In summary, if one would like to adopt these $S$-shaped activation functions, shallower networks are needed if one sets the transition region in these functions to be very narrow. 

\item
The recently more popular ReLU, ELU or swish activation functions behave very differently, as there is not spike, but only a (actual or almost) discontinuity in the Jacobians, so there is no large Frobenius norm at individual $\tilde{J}$ level, just jumps in their entries when the state of the relevant layer changes. Once the local Jacobians multiply into the overall $\mathbb{M}$, the entries in that matrix end up jumping frequently
(multiplication of many Heaviside functions jumping at different values). Therefore, when we scan across all possibilities (varying $i$), the $\mathbb{M}_i$ matrix changes often and $\lambda_i$ tends to explore large ranges, thus have a high probability to hitting large positive values. This effect is likely less pronounced than that of the previous item where large $\lambda_i$ values are more definitively hit (one can alternatively think of these call-option-payoff shaped functions as being less binary so nearby trajectories don't elicit very different activation function outcomes, implying less chaotic propagations), so we predict that ReLU family functions would be more suitable for networks of greater depth.    
\end{itemize}

\subsubsection{Width of layers vs.~connectivity \label{sec:width}}
The width $D$ of the layers of a neural network (number of neurons in each layer, assuming to be a constant for the present discussion for brevity) corresponds to the dimension of the state space of the dynamical system, and $\mathbb{M}$ is a $D\times D$ matrix. How the finite time Lyapunov exponents vary with $D$ is heavily dependent on how the connection matrix $K^{\alpha}_{\beta}$ in Eq.~\eqref{eq:LayerDetails}, between neurons in adjacent layers, changes when we scale up the state space dimension:
\begin{itemize}
\item
First consider increasing $D$ without changing the connectivity between the layers (i.e., the percentage of neurons in the next layer that a particular neuron is connected to, or in other words the percentage of non-vanishing entries in each row of $K^{\alpha}_{\beta}$) or the coupling strength (the typical size of the entries in the weighting matrix $K$), then the sparsity of $\mathbb{M}$ and the typical amplitude of entries in it won't change, resulting in the Frobenius norm scaling as $D^1$ since the number of these entries grow as $D^2$. The number of singular values on the other hand only grows as $D^1$, so the average singular value would have to scale as $\propto \sqrt{D}$. In other words, without changing other features like connectivity and weight ranges etc, a wider neural network tends to possess more positive finite time Lyapunov exponents, thus more likely to stray into the undesirable deep mixing regime (see end of Sec.~\ref{sec:conclusion} for the flip side of this width coin however). Therefore, even without heeding limitations on computational resources, wider networks should be made shallower. 

\item
When we renormalize the coupling strengths so the weighting matrix becomes more like a probability with weights summing up to a constant (e.g., $\sum_{\alpha} K^{\alpha}_{\beta} = 1$), the situation changes. The Frobenius norm now scales as $D^0$, and the average singular value now must scale as $1/\sqrt{D}$, so the dynamics become less chaotic as the width of the layers increases. With this approach, one could thus simultaneous increase both the width and the depth of the network, without degrading performance. Although there doesn't seem to be a strong incentive to doing so, given the higher drain on computational resources.

\item 
Another way to curtail chaos while increasing $D$ is to make the neurons in the wider network more sparsely connected (setting a higher proportion of entries in each row of $K$ to zero). A particularly interesting physical interpretation that is relevant for a subset of such a strategy is related to path dependence, which by definition has a tendency to help retain reliance on initial data, thereby lessen mixing. Given a dynamical system, path dependence can be folded into a particular type of higher dimensionality, with many new auxiliary state variables that only depend on one other variable, since they are supposed to just passively record the past states of that variable and carry them forward without modifications. For example, if $y^{[j+1]\alpha}$ depend not only on $y^{[j]\beta}$ but also on $y^{[j-1]\gamma}$, then we can define an additional set of variables $x^{[j] \beta}$ and simply let them be updated by $x^{[j] \alpha} = y^{[j-1] \alpha}$ (note each $x$ node only depends on one $y$ node in the previous step). This way, the $j+1$th step of the $2D$-dimensional $y\oplus x$ combined system depends only on its step $j$ state, and path dependence is formally removed so a dimensionally expanded version of Eq.~\eqref{eq:dyno} remains valid. 
Mapping such a dynamical system into a neural network, we see that the width of the network can be increased in order to simulate the effect of path dependence, so long as the newly added dimensions don't link up with too many of the existing ones. This could explain why pruning networks sometimes helps when the network is otherwise mixing (suffers from e.g., overfitting problems). 
\end{itemize}

\section{Conclusion \label{sec:conclusion}}
In this brief note, we advocated for the enlistment of ergodic theory to help intuit behaviors of deep neural networks. In particular, we argued that a highly effective deep neural network would likely operate on the edge of chaos. This is the easiest to intuit in the case of a classification problem. The corresponding dynamical system evolution of the neural network rearranges the data into orderly and well-segregated tiles (the specifics of the tiling depends on the hypothesis function choice), each representing a class, and so given any input data to be processed, the output will land in one of them. For this to work, we need the flow to be sufficiently flexible to contort any \emph{contiguous} (the problem needs to be interpolatable for training to be useful) but possibly exotic-looking initial shape representing a class in the initial data space, into a regular simply hypercubic tile (assuming the simplest threshold-based hypothesis function). This means we would need (quasi-)ergodicity, so any initial data point within that shape can eventually arrive at the desired tile site (tiles are larger than infinitesimal neighbourhoods of points, thus strictly speaking we don't need full ergodicity, and subsequently the prefix ``quasi''). On the other hand, we must also avoid the more strongly chaotic mixing behavior, otherwise the class shape being convected by the flow will be shredded and thoroughly mixed up with sibling shapes, making clean well-segregated tiles in output space impossible. In other words, we need moderation in the wildness of the dynamical system trajectories, and just-on-the-cusp of chaotic regime seems ideal. 

We then discussed some network architectural traits that may be advantageous in terms of finding such ergodic-but-not-mixing niche. For our generic deliberation, we have stayed with qualitative properties and intuitive guidelines. However, after the implementation and training of an actual neural network, it should not be prohibitively difficult to compute the actual numerical values of the finite time Lyapunov exponents, as there exists mature and efficient numerical routines for computing singular values. 

The largest values of these could then serve as a quality control indicator that informs on whether the result of the training is suitable for interpolation and extrapolation. This is potentially an important application, as the tunable parameters of a neural network are so very numerous, giving it the ability to always fit to any training data we present to it, but real life applications require the trained network to respond to new situations in a moderate and controlled manner, rather than jerks around wildly (i.e., we want to avoid the overfitting problem). Previous approaches to ensuring that this is the case is largely empirical, by testing the trained network against additional datasets not included in the training stage. This wastes precious labelled data, and one can never be sure these tests properly cover all plausible new situations. The finite time Lyapunov exponents and their ergodic theory significances could possible provide a valuable alternative.  

One could of course also compute the finite time Lyapunov exponents for the purpose of debugging. For example, if it turns out that the trained network lacks expressivity, then one can imagine that the problem might be that the sum $\sum_{\alpha}\lambda^{[N-1]\alpha}_{i}$ is too negative, so the dynamics ends up being highly dissipative and collapses onto a fixed point or an attractor, occupying only a corner of the state space, thereby preventing the network, as an approximator function, from taking up a chunk of the codomain. 

Lastly, we come back to an issue mentioned in Sec.~\ref{sec:fitting}, and briefly speculate on one of the possible reasons why neural networks, with their high dimensions and thus abundance of parameters, do not overfit as one would n\"aively expect. Within our dynamical systems perspective, overfitting, or over-sensitivity to initial data, is a symptom of mixing, which is alternatively characterized as a rapid forgetfulness regarding past history (obviously undesirable for any information processing machine that shouldn't well forget what is being asked of it at the input). As it turns out, having high dimensions (wide networks) might in fact help cure amnesia\footnote{But it may also allow more room for hyper-chaos, so there are two sides to this coin, see Sec.~\ref{sec:width} above for more details on how to suppress the ills.}, and by association overfitting. To see how, recall that a trick for turning non-Markovian systems into a Markovian one is to give past instances of say, a time series, different names, and elevate them into new variables. As a result, a heavily path-dependent system can be re-casted into a high dimensional one (but obviously the reverse recasting is not always possible). Intuitively then, high dimensionality allows for past information to hide and circulate within other dimensions, before eventually being re-injected into the future evolutions of the key variables that we closely track. 
Such a feature, that increases in the degrees of freedom introduces path dependence, is in fact familiar in the context of quantum non-Markovian processes, where past actions, such as interrogations carried out by experimenters, can travel through the surrounding environment (characterized by their vast number of degrees of freedom) and re-assert themselves later on into an open quantum system. For existing literature on this subject, see e.g., the review article \cite{10.3389/frqst.2023.1134583} and references therein. In particular, taking the classical limit of some of the quantum non-Markovianity measures, already proposed, might yield more quantitative barometers useful for guiding neural network designs, should simple trial-and-error tuning, armed with a qualitative understanding as discussed here, prove inadequate.

\acknowledgements
This work is supported by the National Natural Science Foundation of China grants 12073005, 12021003. 

\bibliography{EdgeOfChaos.bbl}

\end{document}